\documentclass[twoside,leqno,twocolumn]{article}

\usepackage[letterpaper]{geometry}

\usepackage{ltexpprt}
\usepackage{hyperref}
\usepackage{amsmath,amssymb,amsfonts}
\usepackage{graphicx}
\usepackage{xcolor}
\usepackage{textcomp}
\usepackage{multicol, multirow}
\usepackage{algorithm}
\usepackage{algpseudocode}
\usepackage{threeparttable}
\usepackage[backend=bibtex]{biblatex}
\bibliography{ref}

\begin{document}

\title{\Large A General-Purpose Transferable Predictor for Neural Architecture Search}

\author{Fred X. Han$^{1}$\thanks{Correspondence to: fred.xuefei.han1@huawei.com}, Keith G. Mills${^2}$, Fabian Chudak$^{1}$, Parsa Riahi${^3}$ \\
Mohammad Salameh$^{1}$, Jialin Zhang${^4}$, Wei Lu$^{1}$, Shangling Jui${^4}$, Di Niu${^2}$\\
\small\baselineskip=9pt $^1$Huawei Technologies Canada, Edmonton,\\
\small\baselineskip=9pt $^2$ University of Alberta,
\small\baselineskip=9pt $^3$ University of British Columbia, \\
\small\baselineskip=9pt $^4$ Huawei Kirin Solution, Shanghai
}

\date{}

\maketitle

% Copyright Statement
% When submitting your final paper to a SIAM proceedings, it is requested that you include
% the appropriate copyright in the footer of the paper.  The copyright added should be
% consistent with the copyright selected on the copyright form submitted with the paper.
% Please note that "20XX" should be changed to the year of the meeting.

% Default Copyright Statement
\fancyfoot[R]{\scriptsize{Copyright \textcopyright\ 2023 by SIAM\\
Unauthorized reproduction of this article is prohibited}}

\begin{abstract}
Understanding and modelling the performance of neural architectures is key to Neural Architecture Search (NAS). Performance predictors have seen widespread use in low-cost NAS and achieve high ranking correlations between predicted and ground truth performance in several NAS benchmarks. However, existing predictors are often designed based on network encodings specific to a predefined search space and are therefore not generalizable to other search spaces or new architecture families. In this paper, we propose a general-purpose neural predictor for NAS that can transfer across search spaces, by representing any given candidate Convolutional Neural Network (CNN) with a Computation Graph (CG) that consists of primitive operators. We further combine our CG network representation with Contrastive Learning (CL) and propose a graph representation learning procedure that leverages the structural information of unlabeled architectures from multiple families to train CG embeddings for our performance predictor. Experimental results on NAS-Bench-101, 201 and 301 demonstrate the efficacy of our scheme as we achieve strong positive Spearman Rank Correlation Coefficient (SRCC) on every search space, outperforming several Zero-Cost Proxies, including Synflow and Jacov, which are also generalizable predictors across search spaces. Moreover, when using our proposed general-purpose predictor in an evolutionary neural architecture search algorithm, we can find high-performance architectures on NAS-Bench-101 and find a MobileNetV3 architecture that attains 79.2\% top-1 accuracy on ImageNet. 
\end{abstract}

\section{Introduction}
\label{sec:intro}

Neural Architecture Search (NAS) automates neural network design and has achieved remarkable performance on many computer vision tasks. A NAS strategy typically performs alternated search and evaluation over candidate networks to maximize a performance metric. While various strategies such as Random Search~\cite{li2019random}, Differentiable Architecture Search~\cite{liu2018DARTS}, Bayesian optimization~\cite{white2019bananas}, and Reinforcement Learning~\cite{pham2018ENAS} can be used for search, architecture evaluation is a key bottleneck to identifying better architectures.

To avoid the excessive cost incurred by training and evaluating each candidate network, most current NAS frameworks resort to performance estimation methods to predict accuracy. Popular methods include 
weight sharing~\cite{pham2018ENAS, liu2018DARTS, cai2019once}, neural predictors~\cite{luo2020semi, white2019bananas} and Zero-Cost Proxies (ZCP)~\cite{abdelfattah2021zero}. The effectiveness of a performance estimation method is mainly determined by the Spearman Rank Correlation Coefficient (SRCC) between predicted performance and the ground truth. A predictor with higher SRCC can better guide a NAS search algorithm 
toward finding superior architectures.

While partial training and weight sharing are used extensively in early NAS works~\cite{pham2018ENAS, zoph2018learning}, thanks to several existing NAS benchmarks that provide an ample amount of labeled networks \cite{dong2020nasbench201,zela2022surrogate}, e.g., NAS-Bench-101~\cite{ying2019bench} offers 423k networks trained on CIFAR-10, there have been many recent developments in training neural predictors for NAS~\cite{wen2020neural,tang2020semi}. As these methods learn to \textit{estimate} performance using labeled architecture representations, they generally enjoy the lowest performance evaluation cost as well as the capacity for continual improvement as more NAS benchmarking data is made available. 

\begin{figure*}[t]
	\centering
	\includegraphics[width=\linewidth]{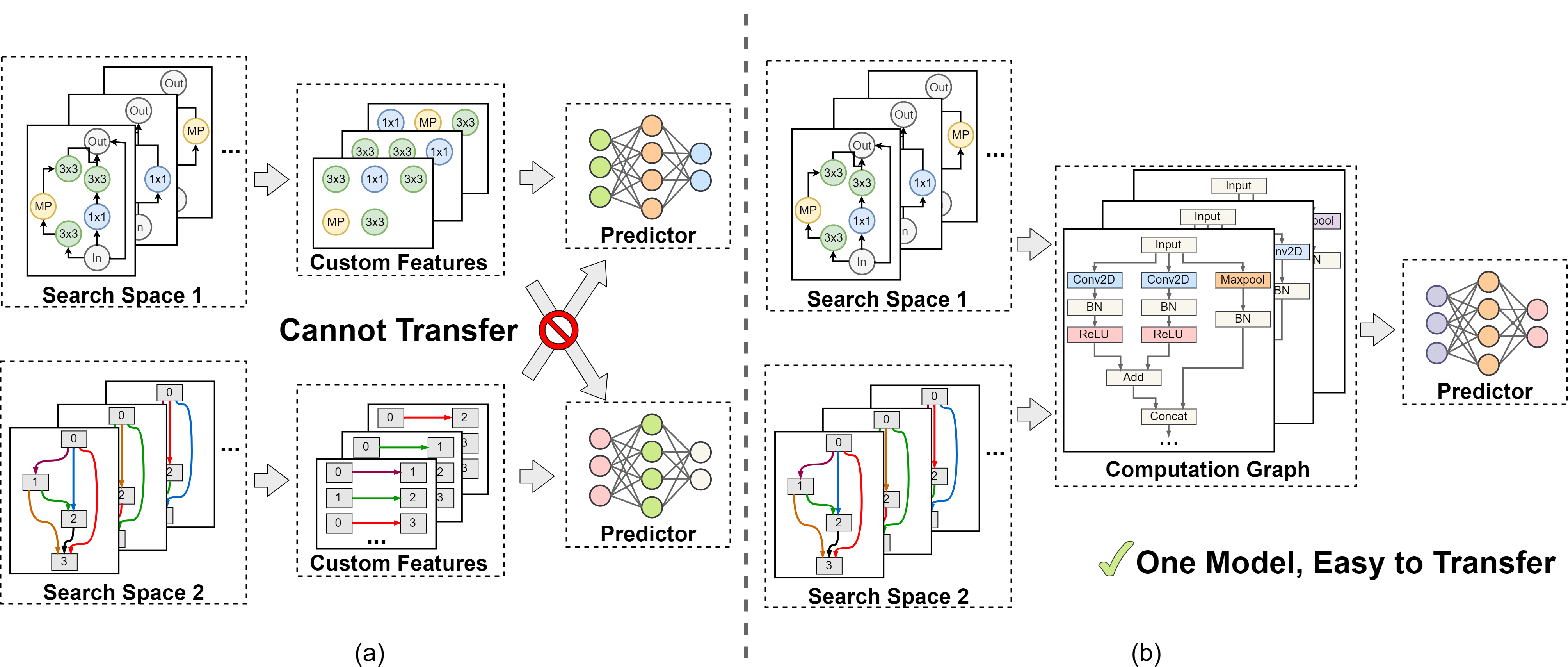}
	\vspace{-5mm}
	\caption{Comparison of architecture representations in neural predictor setups:
	(a) conventional, space-dependent neural predictors, e.g., BANANAS and SemiNAS; (b) our proposed, general-purpose, space-agnostic predictor using Computational Graphs. 
	Best viewed in color.}
	\label{fig:overview}
	\vspace{-3mm}
\end{figure*}

However, a major shortcoming of existing neural predictors is that they are not general-purpose. Each predictor is specialized to process networks confined to a specific search space. For example, NAS-Bench-101 limits its search space to a cell, which is a graph of up to 7 internal operators. Each operator can be one of 3 specific operation sequences. In contrast, NAS-Bench-201~\cite{dong2020nasbench201} and NAS-Bench-301~\cite{zela2022surrogate} adopt different candidate operator sets (with more details in Sec.~\ref{sec:cg}) and network topologies. Thus, a neural predictor for NAS-Bench-101 could not predict the performance for NAS-Bench-201 or NAS-Bench-301 networks. This search-space-specific design severely limits the practicality and transferability of existing predictors for NAS, since for any new search space that may be adopted in reality, a separate predictor must be re-designed and re-trained based on a large number of labeled networks in the new search space.  

Another emerging approach to performance estimation is ZCP methods, which can support any network structure as input. However, the performance of ZCPs may vary significantly depending on the search space. For example, \cite{abdelfattah2021zero} shows that Synflow achieves a high SRCC of 0.74 on the NAS-Bench-201 search space, however the SRCC drops to 0.37 on NAS-Bench-101. Moreover, ZCP methods require the instantiation of a neural network, e.g., performing the forward pass for a batch of training samples, to compute gradient information, and thus incur longer per-network prediction latency. In contrast, neural predictors simply require the network architecture encoding or representation to predict its performance. 

In general, neural predictors offer better estimation quality but do not transfer across search spaces, while ZCPs are naturally universal estimation methods, yet are sub-optimal on certain search spaces. To overcome this dilemma, in this paper, we propose a general-purpose predictor for NAS that is transferable across search spaces like ZCPs, while still preserving the benefits of conventional predictors. Our contributions are summarized as follows:
    
First, we propose the use of \textit{Computation Graphs} to offer a universal representation of neural architectures from different search spaces. Figure~\ref{fig:overview} highlights the differences between (a) existing NAS predictors and (b) the proposed framework. The key is to introduce a universal search space representation consisting of graphs of only primitive operators to model any network structure, such that a general-purpose transferable predictor can be learned based on NAS benchmarks available from multiple search spaces. 

Second, we propose a framework to learn a \textit{generalizable} neural predictor by combining recent advances in Graph Neural Networks (GNN)~\cite{morris2019weisfeiler} and Contrastive Learning (CL)~\cite{chen2020simple}. Specifically, we introduce a graph representation learning process to learn a generalizable architecture encoder based on the structural information of vast unlabeled networks in the target search space. The embeddings obtained this way are then fed into a neural predictor, which is trained based on labeled architectures in the source families, achieving transferability to the target search space. 
    
Experimental results on NAS-Bench-101, 201 and 301 show that our predictor can obtain high SRCC on each search space when fine-tuning on no more than 50 labeled architectures in a given target family. Specifically, we outperform several ZCP methods like Synflow and obtain SRCC of 0.917 and 0.892 on NAS-Bench-201 and NAS-Bench-301, respectively. Moreover, we use our predictor for NAS with a simple evolutionary search algorithm and have found a high-performance architecture (with 94.23\% accuracy) in NAS-Bench-101 at a low cost of 700 queries to the benchmark, outperforming other \textit{non-transferable} neural predictors such as SemiNAS~\cite{luo2020semi} and BANANAS~\cite{white2019bananas}. Finally, we further apply our scheme to search for an ImageNet~\cite{deng2009imagenet} network and have found a Once-for-All MobileNetV3 (OFA-MBv3) architecture that obtains a top-1 accuracy of 79.2\% which outperforms the original OFA. 
\section{Related Work}
\label{sec:related}

Neural predictors are a popular choice for performance estimation in low-cost NAS. Existing predictor-based NAS works include SemiNAS~\cite{luo2020semi}, which adopts an encoder-decoder setup for architecture encoding/generation, and a simple neural performance predictor that predicts based on the encoder outputs. SemiNAS progressively updates an accuracy predictor during the search. BANANAS~\cite{white2019bananas} relies on an ensemble of accuracy predictors as the inference model in its Bayesian Optimization process. \cite{tang2020semi} construct a similar auto-encoder-based predictor framework and supplies additional unlabeled networks to the encoder to achieve semi-supervised learning. \cite{wen2020neural} propose a sample-efficient search procedure with customized predictor designs for NAS-Bench-101 and ImageNet~\cite{deng2009imagenet} search spaces. NPENAS \cite{wei2022npenas}, BRP-NAS \cite{dudziak2020brp} are also notable predictor-based NAS approaches. By contrast, our approach pre-trains architecture embeddings that are not restricted to a specific underlying search space. 

Zero-Cost Proxies (ZCP) are originally proposed as parameter saliency metrics in model pruning techniques. With the recent advances of pruning-at-initialization algorithms, only a single forward/backward propagation pass is needed to assess the saliency. Metrics used in these algorithms are becoming an emerging trend for transferable, low-cost performance estimation in NAS. \cite{abdelfattah2021zero} transfers several ZCP to NAS, such as Synflow~\cite{tanaka2020pruning}, Snip~\cite{lee2018snip}, Grasp~\cite{wang2020picking} and Fisher~\cite{turner2019blockswap}. Zero-Cost Proxies could work on any search space and \cite{abdelfattah2021zero} shows that on certain search spaces, they help to achieve results that are comparable to predictor-based NAS algorithms. However, the performance of ZCPs are generally not consistent across different search spaces. Unlike neural predictors such as ours, they require instantation of candidate networks as well as sample data to compute gradient metrics. 
\section{A Unified Architecture Representation}
\label{sec:cg}

In this section, we discuss how to represent neural networks. First, we elaborate on \textit{operator-grouping} within NAS, how it simplifies architecture representation while hindering transferability between search spaces. Then, we introduce our Computational Graph (CG) framework and how it solves the transferability problem. 

\subsection{Abstract Operation Representations}
Without the loss of generality, we consider operations in neural networks and define a \textit{primitive} operator as an atomic node of computation. In other words, a primitive operator is one like Convolution, ReLU, Add, Pooling or Batch Normalization (BN), which are single points of execution that cannot be further reduced. 

\begin{figure*}[t]
	\centering
	\includegraphics[width=\linewidth]{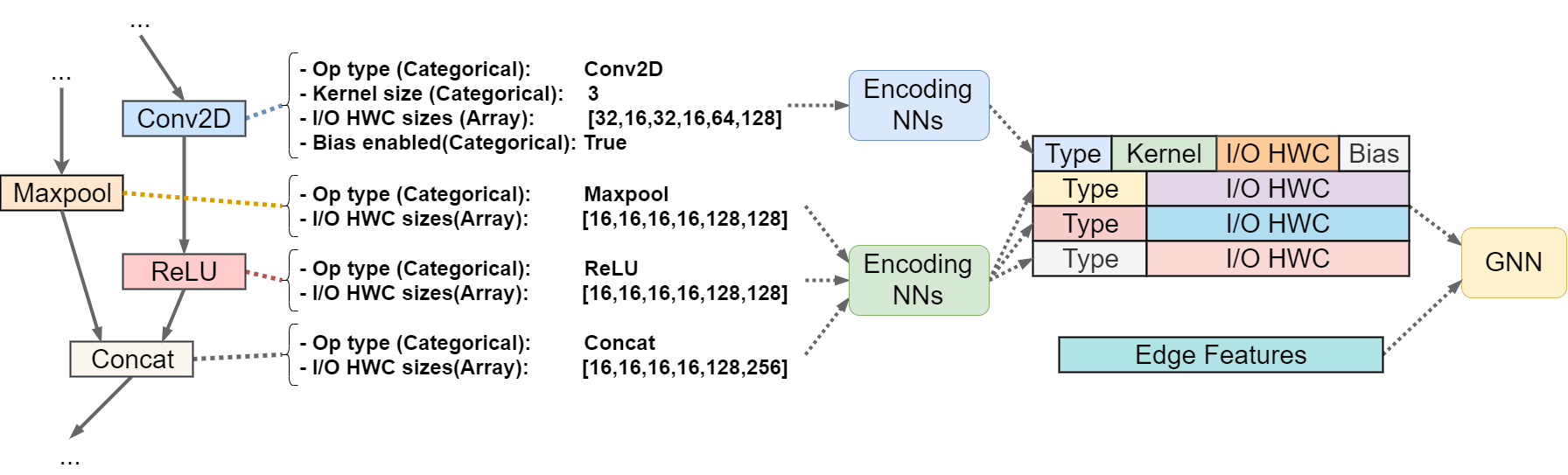}
	\caption{An example illustrating the key graphical features extracted from compute graphs and how we encode them as GNN node features. All nodes contain input and output tensor HWC sizes. Nodes with trainable weights contain additional features on the weight matrix dimensions and bias.}
	\label{fig:cg_feature}
\end{figure*}

\textit{Operator-grouping} is an implicit but widely adopted technique for improving the efficiency of NAS. It is a coarse abstraction where atomic operations are grouped and re-labeled according to pre-defined sequences. For example, convolution operations are typically grouped with BN and ReLU~\cite{cai2019once}. Rather than explicitly representing all three primitive operations independently, a simpler representation describes all three primitives as one grouped operation pattern that is consistent throughout a search space. We can then use these representations as feature inputs to a search algorithm~\cite{rezaei2021generative, luo2020semi} or neural predictor~\cite{wen2020neural, dudziak2020brp}, as prior methods do. 

While operator grouping provides a useful method to abstract how we represent neural networks, it hinders transferability as groupings may not be consistent across search spaces. Take the aforementioned convolution example: Although convolutions are typically paired with BN and ReLU operations, the ordering of these primitives can vary. While NAS-Bench-101 use the `Conv-BN-ReLU' ordering, NAS-Bench-201 use `ReLU-Conv-BN', thus forming a discrepancy that current neural predictors do not take into account. 

Compounding this issue, the set of operations can differ by search space. While NAS-Bench-101 considers Max Pooling, NAS-Bench-201 only uses Average Pooling, and NAS-Bench-301 uses complex convolutions with larger kernel sizes not found in either 101 nor 201. A neural predictor trained on any one of these search spaces using operator-grouping could not easily transfer to another. Thus, operator-grouping is the main culprit of low transferability in existing predictors. We provide a table enumerating the operator groupings for each search space in the supplementary materials. 

\subsection{Computational Graphs}
To construct a neural predictor that is transferable between search spaces, we consider a representation that can \textit{generalize} across multiple search spaces. We define a Computation Graph (CG) as a detailed representation of a neural network without any customized grouping, i.e., each node in the graph is a primitive operator and the network CG is made up of only such nodes and edges that direct the flow of information. Without operator-grouping, CGs define a search space that could represent any network structure since the number of primitive operators is usually far less than the number of possible groupings.

While there are many potential ways to construct a CG, we adopt a simple approach of using the model optimization graph maintained by deep learning frameworks like TensorFlow \cite{tensorflow2015-whitepaper} or PyTorch \cite{NEURIPS2019_9015}. As the name suggests, a model optimization graph is originally intended for gradient calculations and weight updates, and it is capable of supporting any network structure. In this work, all the CGs used in our experiments are simplified from TensorFlow model optimization graphs. Specifically, we extract the following nodes from a model optimization graph to form a CG: 
\begin{itemize}
    \item Nodes that refer to trainable neural network weights. For these nodes, we extract the atomic operator type, such as Conv1D, Conv2D, Linear, etc., input/output channel sizes, input/output image height/width sizes, weight matrix dimensions (e.g., convolution kernel tensor) and bias information as node features. 
    \item Nodes that refer to activation functions like ReLU or Sigmoid, pooling layers like Max or Average, as well as Batch Normalization. For these nodes, we extract the operator type, input/output channel and image height/width sizes.
    \item Key supplementary nodes that indicate how information is processed, e.g., addition, concatentation and element-wise multiplication. For these nodes, we also extract the type, input/output channel and image height/width sizes.
\end{itemize}
Figure~\ref{fig:cg_feature} illustrates how we transform a computation graph into learnable feature vectors. Formally, a computation graph $G$ consists of a vertex set $V=\{v_1, v_2, v_3, ...\}$ and an edge set $E$, where $v$ refers to a primitive operator node and $E$ contains pairs of vertices $(v_s, v_d)$ indicating a connection between $v_s$ and $v_d$. Under this definition, the problem of performance estimation becomes finding a function $F$, e.g., a Graph Neural Network (GNN), such that for computation graph $G_i$, which is generated from a candidate neural network, $F(G_i)=Y_i$, where $Y_i$ is the ground truth test score. 

Representing networks as CGs enable us to break the barrier imposed by search space definitions and fully utilize all available data for predictor training, regardless of where a labeled network is from, e.g., NAS-Bench-101, 201 or 301. We could also effortlessly transfer a predictor trained on one search space to another, by simply fine-tuning it on additional data from the target space. 
\section{Neural Predictor via Graph Representation Learning}

In this section, we propose a two-stage approach to improve generalization and leverage unlabeled data via Contrastive Learning (CL). We first find a vector representation, i.e., graph embedding, which converts graph features with a variable number of nodes into a fixed-sized latent vector via a graph CL procedure, before feeding the latent vector to an MLP accuracy regressor. Given a target family for performance estimation, a salient advantage of our approach is its ability to leverage unlabeled data, e.g., computation graphs of the target family, which are typically available in abundance. In fact, our approach is able to jointly leverage labeled and unlabeled architectures from multiple search spaces to maximally utilize available information.

For each CG $G_i$, we would like to produce a vector representation $h_i \in \mathbb{R}^e$ for a fixed hyper-parameter $e$. We would like to infer relationships between networks by considering the angles between vector representations. Our CL-based approach learns representations where only \textit{similar} CGs have close vector representations. 

\subsection{Contrastive Learning Frameworks}
 
SimCLR~\cite{chen2020simple,chen2020big} and SupCon~\cite{khosla2020supervised} apply CL to image classification. The general idea is to learn a base encoder to create vector representations $h$ of images. To train the base encoder, a projection head $Proj(*)$ maps the vector representations into a lower-dimensional space $z \in \mathbb{R}^p; p < e$ to optimize a \textit{contrastive loss}. 

The contrastive loss forces representations of similar objects to be in agreement. Consider a batch of $N$ images, whose vector representation is $I=\{h_1,h_2,\ldots,h_N\} \subset \mathbb{R}^e$. For each $G_i$, let $z_i=Proj(h_i)\in \{||z||=1: z \in \mathbb{R}^p\}$, and let the cosine similarity be $sim(z_i,z_j)=z_i\cdot z_j/\tau$, where the temperature $\tau>0$ and $\cdot$ is the dot product. The agreement $\chi_{i, j}$ between two arbitrary indices $i$ and $j$ is given by 

\begin{equation}
    \centering
    \label{eq:agreement}
    \chi_{i, j} = \log \frac{\exp(sim(z_i,z_j))}{\sum_{r\not=i} \exp(sim(z_i,z_r))}. 
\end{equation}
A primary distinction between SimCLR and SupCon is how we determine if two different objects are similar. 

SimCLR considers the unsupersived context where we do not have access to label/class information. Data augmentation plays a crucial role. For each \textit{anchor} index $i$ in a batch, we apply a transform or slight perturbation to create an associated \textit{positive} element $j(i)$, while all other samples $r\not=i,j(i)$ are \textit{negative} indices. The SimCLR loss function, 

\begin{equation}
    \centering
    \label{eq:simclr}
    \mathcal{L}_{SimCLR} = - \sum_{i\in I} \chi_{i, j(i)}, 
\end{equation}
serves to maximize agreement between the original and augmented images. 

\label{sec:cl-pred}
\begin{figure}[t]
	\centering
	\includegraphics[width=\linewidth]{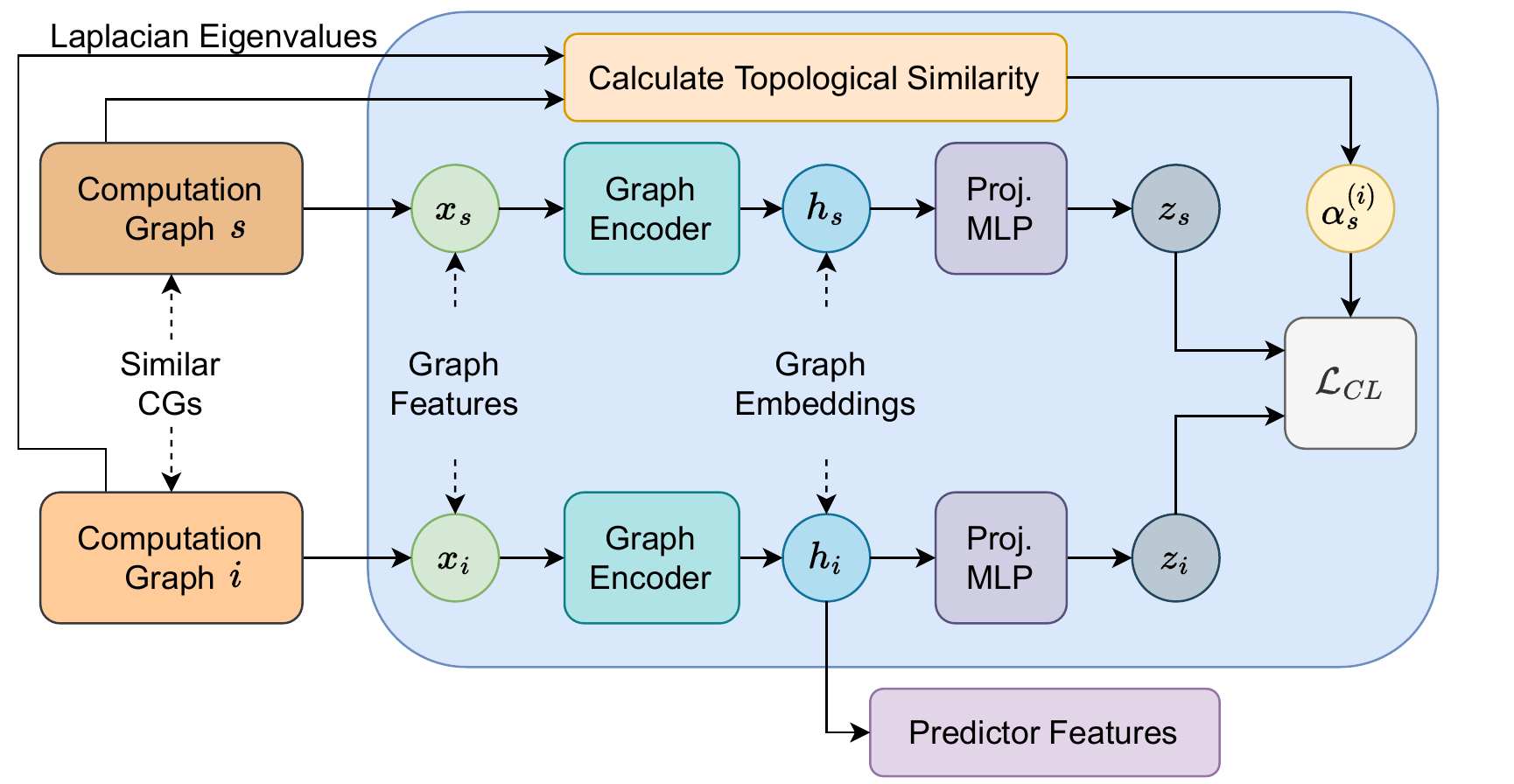}
	\caption{Contrastive Learning framework. We apply a graph encoder to produce embeddings. Next, a projection layer produces representation vectors whose similarity we compare using the CL loss, weighted according to the Laplacian Eigenvalues of each CG.}
	\label{fig:cl}
\end{figure}

By contrast, if classes are known, we can use the the SupCon loss function~\cite{khosla2020supervised}, 
\begin{equation}
\centering
\label{eq:supcon}
\mathcal{L}_{SupCon} = \sum_{i\in I} \frac{-1}{|P(i)|} \sum_{s \in P(i)} \chi_{i, s},
\end{equation}
where $P(i)$ is the set of non-anchor indices whose class is the same as $i$. Thus, not just $j(i)$, but all indices whose class is the same as $i$ contribute to the probability of positive pairs. 

Next, we construct a contrastive loss for CGs. We consider the challenges and advantages of using graphs as data as well as the overall problem of regression instead of classification.

\subsection{Computational Graph Encodings}

\begin{table*}[t]
\caption{Spearman correlation coefficients ($\rho$). We compare our CL encoding scheme to a GNN encoder as well as several ZCPs. For our CL and GNN, we report the mean and standard deviation over 5 fine-tuning runs. }
\label{tab-corr}
\small
\begin{center}
\scalebox{0.9}{
\begin{tabular}{|l|c|c|c|} \hline
\textbf{Method}                              & \textbf{NAS-Bench-101} & \textbf{NAS-Bench-201}  & \textbf{NAS-Bench-301} \\ 
\hline
Synflow \cite{tanaka2020pruning}    & 0.361  & 0.823    & -0.210\\
Jacov \cite{mellor2021neural}       & 0.358  & 0.859    & -0.190\\
Fisher \cite{turner2019blockswap}   & -0.277 & 0.687 & -0.305\\
GradNorm \cite{abdelfattah2021zero} & -0.256 & 0.714   & -0.339\\
Grasp \cite{wang2020picking}        & 0.245  & 0.637    & -0.055\\
Snip \cite{lee2018snip}             & -0.165 & 0.718   & -0.336\\
\hline
GNN-fine-tune                        & 0.542 $\pm$ 0.14 & 0.884 $\pm$ 0.03  & 0.872 $\pm$ 0.01\\
CL-fine-tune                         & \textbf{0.553} $\pm$ 0.09 & \textbf{0.917} $\pm$ 0.01  & \textbf{0.892} $\pm$ 0.01 \\
\hline
\end{tabular}
}
\end{center}
\vspace{-5mm}
\end{table*}

Figure~\ref{fig:cl} provides a high-level overview of our scheme. When applying CL to CGs, we start by considering the similarity between CGs. We leverage the rich structural information CGs provide by encoding each atomic primitive within a network as a node. Specifically, our approach uses spectral properties of undirected graphs~\cite{dwivedi2021generalization}. Given a CG with $|V|$ nodes, we consider its underlying undirected graph $G'$. Let $A \in \{0,1\}^{|V| \times |V|}$ be its adjacency matrix and $D\in \mathbb{Z}^{|V| \times |V|}$ be its degree diagonal matrix. The normalized Laplacian matrix is defined as 

\begin{equation}
    \centering
    \label{eq:graph_laplacian}
    \Delta  = I -D^{-1/2}AD^{-1/2} = U^T\Lambda U ,
\end{equation}
where $\Lambda \in \mathbb{R}^{|V| \times |V|}$ is the diagonal matrix of eigenvalues and $U$ the matrix of eigenvectors. Eigenvalues $\Lambda$ encode important connectivity features. For instance, 0 is the smallest eigenvalue and has multiplicity 1 if and only if $G'$ is connected. Smaller eigenvalues focus on general features of the graph, whereas larger eigenvalues focus on features at higher granularity; we refer the reader to \cite{wills2019metrics} for more details. 

More generally, we can use the eigenvalues of $\Delta$ to measure \textit{pseudo-distance} between graphs. Given two CGs $g_1,g_2$, we compute the spectral distance $\sigma_S(g_1,g_2)$ as the Euclidean norm of the corresponding $k=11$ smallest eigenvalues. 

Our constrastive loss incorporates elements from both SimCLR and SupCon in addition to spectral distance. As our task of interest is regression rather than classification, we replace the positive-negative binary relationship between samples in a batch with a probability distribution over all pairs which smoothly favors similar computation graphs. 

First, if the family of networks each CG belongs to is known, e.g., NAS-Bench-101, 201, etc., we can treat the family affiliation as a class and follow the SupCon approach. Second, rather than using the uniform distribution over $P(i)$ as in Equation~\ref{eq:supcon}, we use a convex combination over $P(i)$ based on the similarity of the corresponding computation graphs. Overall, our loss function is given as 
\begin{equation}
    \centering
    \label{eq:cl_loss}
    \mathcal{L}_{CL} = - \sum_{i\in I} \sum_{s \in P(i)} \alpha_s^{(i)} \chi_{i, s}, 
\end{equation}
where $\alpha_s^{(i)} \ge 0$ and $\sum_{s \in P(i)} \alpha_s^{(i)}=1$. For computation graph $i$, we simply define $\alpha_*^{(i)}$ to be the softmax of $\sigma_S(i,*)$ with temperature 0.05. 

Finally, there are challenges associated with data augmentation for CGs. Slightly perturbing a CG may drastically change its accuracy on a benchmark, e.g., changing an activation function or convolution~\cite{mills2021profiling}. In more severe scenarios, arbitrary small changes to a computation graph may make it fall outside of the family of networks of interest or even result in a graph that does not represent a functional neural network at all. To address this, rather than randomly perturbing a CG, we use $\sigma$ to randomly pick a very similar graph from the training set to form a positive pair. As suggested in \cite{khosla2020supervised}, we learn the embeddings using large batch sizes. We enumerate the structure of our predictor and other details in the supplementary materials. 
\section{Experimentation}
\label{sec:exp}

\begin{table*}[t!]
\caption{Search results on NAS-Bench-101, 201 and 301 using the same EA search algorithm but with different performance estimation methods. \#Q represents the number of unique networks queried during search. Note that the \#Q for CL-fine-tune also counts the fine-tuning instances.}
\label{tab-search1}
\begin{center}
\scalebox{0.9}{
\begin{tabular}{|l|l|l|l|l|l|l|l|l|} \hline
   \multirow{2}{*}{\textbf{Method}}                 & \multicolumn{3}{c|}{\textbf{NAS-Bench-101}} & \multicolumn{3}{c|}{\textbf{NAS-Bench-201}} & \multicolumn{2}{c|}{\textbf{NAS-Bench-301\tnote{*}}} \\\cline{2-9}
                   & \#Q & Acc. (\%) & Rank   & \#Q & Acc. (\%) & Rank  &  \#Q & Acc. (\%)\\ 
\hline
Random             & 700 & 94.11 $\pm$ 0.10 & 26.0    & 90 & 93.91 $\pm$ 0.2 & 104                         & 800 & 94.75 $\pm$ 0.08 \\
Synflow            & 700 & 94.18 $\pm$ 0.05 & 5.8     & 90 & \textbf{94.37} $\pm$ 0.0 & \textbf{1.0}                            & 800 & 94.60 $\pm$ 0.11\\
\hline
CL-fine-tune       & 700 & \textbf{94.23} $\pm$ 0.01 & \textbf{2.2}    & 90 & \textbf{94.37} $\pm$ 0.0 & \textbf{1.0}           & 800 & \textbf{94.83} $\pm$ 0.06\\
\hline
\end{tabular}
}
\end{center}
\vspace{-3mm}
\end{table*}

In this section, we evaluate our proposed transferable predictor by comparing ranking correlations with other transferable Zero-Cost Proxies (ZCP) on popular NAS benchmarks. Then, we compare with other neural predictors in the literature by performing search, showing that methods with higher ranking correlations often produce better results. 

\subsection{Comparison of ranking correlations}
\label{sec:exp-corr}

We consider the search spaces of NAS-Bench-101~\cite{ying2019bench}, NAS-Bench-201~\cite{dong2020nasbench201} and NAS-Bench-301~\cite{zela2022surrogate}, with 50k, 4096\footnote{CIFAR-10 architectures that do not contain `zeroize'.} and 10k overall CG samples, respectively. We differentiate between \textit{target} and \textit{source} families depending on our configuration. We treat target families as unseen test domains and assume we only have access to a limited amount of labeled target data, yet a large amount of unlabeled target CGs. We use source families to train our predictors and we assume labels are known for each CG. 

We consider the three cases where one of NAS-Bench-101, 201 or 301 is the held-out target family, and use the other two as source families. We use structural information from unlabeled samples in the target family to train our CL encoder in an unsupervised manner. Then, we use labeled data from the source families to train an MLP predictor using supervised regression. Finally, we use a small amount of labeled data from the target family to fine-tune the MLP predictor. 

In addition to our CL-based encoder and ZCPs, we consider a simple GNN~\cite{morris2019weisfeiler} regressor baseline that we can train and fine-tune in an end-to-end fashion. We provide implementation details for each predictor in the supplementary materials. 

We sample 5k instances from NAS-Bench-101, 4096 instances from NAS-Bench-201 and 1k instances from NAS-Bench-301 when they are the target test family, and use all available data when they are a source family. When fine-tuning, we use 50 CGs from NAS-Bench-101 and NAS-Bench-301 and 40 CGs if NAS-Bench-201 is the target family. We execute the ZCPs on the test sets and report the Spearmans Rank Correlation Coefficient (SRCC) 
($\rho \in [-1, 1]$). SRCC values closer to 1 indicate higher ranking correlations between predictions and the ground truth accuracy.

Table~\ref{tab-corr} summarizes the results. First, we note that our CL-based encoder achieves the best SRCC in all three target family scenarios. On NAS-Bench-101, only our CG-based schemes can achieve SRCC above 0.5, while some ZCPs fail to even achieve positive correlation. On NAS-Bench-201, only our CL scheme is able to achieve over 0.9 SRCC while the GNN baseline and the best ZCP methods, Synflow and Jacov, achieve around 0.85 SRCC. Similar to \cite{abdelfattah2021zero}, on NAS-Bench-301, all of the ZCP schemes fail to achieve positive ranking correlation. By contrast, both of our predictors achieve over 0.85 SRCC on 301. Moreover, the CL encoder with fine-tuning achieves very low standard deviation on all three benchmarks. This indicates stable, consistent performance. Overall, our findings demonstrate the utility of CGs as generalizable, robust neural network representations and the capacity of our CL scheme to learn rich graph features through unlabeled data.

\subsection{Search results}
\label{sec:exp-search}

We now demonstrate that our predictor is a superior choice for transferable performance estimation in NAS. We vary the performance estimator and report the accuracy and rank of the best architecture found. 
Specifically, we compare the CL-fine-tune predictor for a given target family against the Synflow ZCP and a random estimation baseline. We pair each predictor with a simple evolutionary approach that we detail in the supplementary materials. 

For each method, we conduct 5 search runs on NAS-Bench-101 which has 423,624 labeled candidates, NAS-Bench-201 with 15,625 searchable candidates, and NAS-Bench-301 with over $10^{18}$ candidates. To establish a fair comparison, we query (\#Q) the same number of networks. Intuitively, the number of queries made to the benchmark simulates the real-world cost of performance evaluation in NAS. 

Table~\ref{tab-search1} reports our results. On NAS-Bench-101 and 301, our CL-fine-tune predictors consistently find better architectures than either Synflow or the random estimation baseline. On average, our CL predictor can find the 2nd best NAS-Bench-101 architecture while Synflow finds the 6th best, and does not come close in terms of performance on NAS-Bench-301. On the smallest search space, NAS-Bench-201, both our CL-fine-tune predictor and Synflow easily and consistently find the best CIFAR-10 architecture. We recall from Table~\ref{tab-corr} how none of the ZCP methods could achieve positive SRCC on NAS-Bench-301 and note how poor estimation performance reflects downstream search results as a random estimation baseline outperforms Synflow on that search space. By contrast, with a small amount of fine-tuning data, our CL-fine-tune predictor achieves exceptional transferability across architecture families and is able to better guide the search process.

\begin{table}[t]
\caption{Search performance of our CL-fine-tuning predictor and EA search algorithm against other NAS approaches on NAS-Bench-101. We report the best architecture accuracy and number of queries.} 
\label{tab-search2}
\begin{center}
\scalebox{0.9}{
\begin{tabular}{|l|c|c|} \hline
\textbf{NAS algorithm}                      & \textbf{\#Queries} & \textbf{Best Acc.} \\
\hline
Random Search                               & 2000 & 93.66\%   \\
SemiNAS \cite{luo2020semi}                 & 2000 & 94.02\% \\
SemiNAS (RE) \cite{luo2020semi}            & 2000 & 94.03\% \\
SemiNAS (RE) \cite{luo2020semi}            & 1000 & 93.97\% \\
BANANAS \cite{white2019bananas}            & 800 & \textbf{94.23}\%  \\
GA-NAS \cite{rezaei2021generative}         & 378 & \textbf{94.23}\%  \\
Neural-Predictor-NAS \cite{wen2020neural}  & 256 & 94.17\% \\
NPENAS \cite{wei2022npenas}                & 150 & 94.14\% \\
BRP-NAS \cite{dudziak2020brp}              & 140 & 94.22\% \\
\hline
\textbf{EA + CL-fine-tune}                           & 700 & \textbf{94.23}\%  \\
\hline
\end{tabular}
}
\end{center}
\vspace{-3mm}
\end{table}

Next, we compare our best search results on NAS-Bench-101 to other state-of-art NAS approaches in Table~\ref{tab-search2}. We observe that our EA + CL-fine-tune setup is competitive with other NAS algorithms. For instance, our setup requires fewer queries to find the second-best architecture (94.23\%) in NAS-Bench-101 compared to BANANAS. The only other scheme which achieves that level of performance is our previous work, GA-NAS~\cite{rezaei2021generative}. Moreover, our search result is better than SemiNAS in terms of the number of queries and accuracy. It is also critical to note that, due to the generalizable structure of our CGs, we can transfer our predictor to other search spaces with only a small amount of labeled data for fine-tuning. This is a unique advantage among neural predictors. 

Finally, we further test the efficacy of our predictor in NAS by searching on the large-scale classification dataset ImageNet~\cite{deng2009imagenet}. To reduce carbon footprint, we search on the Once-for-All (OFA)~\cite{cai2019once} design space for MobileNetV3 (MBv3)~\cite{howard2019searching}, which allows us to quickly query the accuracy of a found network using a pre-trained supernet. We train a CL encoder and predictor on NAS-Bench-101, 201 and 301, then fine-tune on 50 random networks from the OFA-MBv3 search space. 

Table~\ref{tab-imagenet} reports our findings. Using the CL predictor, we are able to find an architecture that achieves over 79.0\% top-1 accuracy on ImageNet, outperforming works on the same search space such as the original OFA~\cite{cai2019once} and \cite{mills2021profiling}. These results further reinforce the generalizability of our scheme, as we are able to train a neural predictor on three CIFAR-10 benchmark families, then then transfer it to perform NAS on a MobileNet family designed for ImageNet.

\begin{table}
\caption{ImageNet search results. We search in the OFA-MBv3 space and report the top-1 accuracy acquired using the OFA-MBv3 supernet. Most baseline results are provided by \cite{cai2019once}.}
\label{tab-imagenet}
\begin{center}
\scalebox{0.9}{
\begin{tabular}{|l|c|}
\hline
\textbf{Model}                                           &\textbf{ Top-1 Acc. (\%)}      \\
\hline
MobileNetV2 \cite{sandler2018mobilenetv2}         & 72.0                 \\
MobileNetV2 \#1200 \cite{sandler2018mobilenetv2}  & 73.5                 \\
MobileNetV3-Large \cite{howard2019searching}      & 75.2                 \\
NASNet-A \cite{zoph2018learning}                  & 74.0                 \\
DARTS \cite{liu2018DARTS}                         & 73.1                 \\
SinglePathNAS \cite{guo2020single}                & 74.7                 \\
OFA-Large \cite{cai2019once}                      & 79.0                 \\
OFA-Base \cite{mills2021profiling}                & 78.9                 \\
\hline
\textbf{OFA-MBv3-CL}                              & \textbf{79.2}        \\              
\hline
\end{tabular}
}
\end{center}
\end{table}
\section{Conclusion}
\label{sec:conclusion}
In this work, we propose the use of Computational Graphs (CG) as a universal representation for CNN structures. On top of this representation, we design a novel, well-performing, transferable neural predictor that incorporates Contrastive Learning to learn robust embeddings for a downstream performance predictor. Our transferable predictor alleviates the need to manually design and re-train new performance predictors for any new search spaces in the future, which helps to further reduce the computational cost and carbon footprint of NAS. Experiments on the NAS-Bench-101, 201 and 301 search spaces verify our claims. Results show that our predictor is superior to many Zero-Cost Proxy methods on these search spaces. By pairing our predictor with an evolutionary search algorithm we can find a NAS-Bench-101 architecture that obtains 94.23\% CIFAR-10 accuracy and a MobileNetV3 architecture that attains 79.2\% ImageNet top-1 accuracy.

\printbibliography

\clearpage
\newpage
\section{Supplementary Materials}
\label{sec:supp}

\begin{table}[t]
    \caption{Candidate operation groupings for NAS-Bench-101, 201 and 301. 
    We report the sequence of atomic operations in each grouping as well as the number of nodes we would use to represent it as a CG subgraph. `BN' means Batch Normalization.}
    \label{tab:op_list}
    \begin{center}
    \scalebox{0.9}{
    \begin{threeparttable}
    \begin{tabular}{|c|c|c|} \hline
    \textbf{Op. Name} & \textbf{Sequence} & \textbf{\#Nodes} \\ \hline
    \multicolumn{3}{|c|}{NAS-Bench-101~\cite{ying2019bench}} \\ \hline 
    1 $\times$ 1 Conv & [Conv, BN, ReLU] & 3 \\ 
    3 $\times$ 3 Conv & [Conv, BN, ReLU] & 3 \\ 
    3 $\times$ 3 Max Pool & [Max Pool] & 1 \\ \hline 
    \multicolumn{3}{|c|}{NAS-Bench-201~\cite{dong2020nasbench201}} \\ \hline 
    Zeroize & -- & 0 \\ 
    Skip-Connect & [Identity] & 0 \\ 
    1 $\times$ 1 Conv & [ReLU, Conv, BN] & 3 \\ 
    3 $\times$ 3 Conv & [ReLU, Conv, BN] & 3 \\ 
    3 $\times$ 3 Ave. Pool & [Ave. Pool] & 1 \\ \hline 
    \multicolumn{3}{|c|}{NAS-Bench-301~\cite{zela2022surrogate, liu2018DARTS}} \\ \hline 
    Zeroize & -- & 0 \\ 
    Skip-Connect & [Identity] & 0 \\ 
    3 $\times$ 3 Sep. Conv\tnote{*} & [ReLU, Conv, Conv, BN & 8 \\
     & ReLU, Conv, Conv, BN] & \\
    5 $\times$ 5 Sep. Conv\tnote{*} & [ReLU, Conv, Conv, BN & 8 \\
     & ReLU, Conv, Conv, BN] & \\ 
    3 $\times$ 3 Dil. Conv\tnote{$\dagger$} & [ReLU, Conv, Conv, BN] & 4 \\
    5 $\times$ 5 Dil. Conv\tnote{$\dagger$} & [ReLU, Conv, Conv, BN] & 4 \\
    3 $\times$ 3 Ave. Pool & [Ave. Pool] & 1 \\ 
    3 $\times$ 3 Max Pool & [Max Pool] & 1 \\ \hline
    \end{tabular}
    \begin{tablenotes}
        \item{*} Depthwise Separable Convolution~\cite{sandler2018mobilenetv2}. The first convolution in a `Conv, Conv' pair has `groups' equal to 
        input channels, and the second conv 
        is 1 $\times$ 1.
        \item{$\dagger$} Dilation (Atrous) Convolution~\cite{chen2017deeplab}. The first conv 
        in a pair has a dilated kernel, while the second is 1 $\times$ 1.
    \end{tablenotes}
    \end{threeparttable}
    }
    \end{center}
    \vspace{-6mm}
\end{table}

\subsection{Architecture Groupings}
Table~\ref{tab:op_list} enumerates the operator groupings for NAS-Bench-101, NAS-Bench-201 and NAS-Bench-301, respectively. Furthermore, we list the number of nodes a Computational Graph (CG) needs to encode each sequence. Finally, we note the presence of depthwise separable~\cite{sandler2018mobilenetv2} and dilated (atrous) convolutions~\cite{chen2017deeplab} in NAS-Bench-301, in contrast to the regular convolutions found in NAS-Bench-101 and NAS-Bench-201.

\subsection{Evolutionary Search}
\label{sec:search}

\begin{algorithm}[t]
	\caption{Evolutionary Search Algorithm (EA)} 
	\label{alg:ea}
	\begin{algorithmic}[1]
	    \State \textbf{Input:} Random architecture set $P$; Predictor $M$; Budget $B$; NAS-Benchmark $D$; 
	    \For {$t=1,2,\ldots, T$}
	        \State $P_{best} \leftarrow$ top-$k(P)$
	        \State $P_{new} \leftarrow$ Mutation\_and\_Crossover($P_{best}$)
	        \State $P_{new} \leftarrow$ Rank($P_{new}, M$) \Comment{Sort using predictor}
	        \State $P_{child} = \emptyset$
	        \For {$b=1,2,\ldots, B$}
	            \State $P_{child} \leftarrow P_{child}$ + Query($P_{new}[b], D$)
	        \EndFor
	        \State $P$ $\leftarrow$ $P$ + $P_{child}$
	    \EndFor
	\end{algorithmic} 
\end{algorithm}

To perform search, 
we adopt a common Evolutionary Algorithm (EA) 
that consists of 
a combination of crossover and mutation procedures to create a larger pool of 
child architectures.
We provide a high-level description for our procedure. 

Given a population of architectures $P$, at each iteration $t=1,2,\ldots, T$, we select the current top-$k$ best architectures as potential parents and denote this subset as $P_{best}$. To perform \textit{crossover}, we select two architectures $P_{best}$, and denote them as ${parent}_1$ and ${parent}_2$. Then, we randomly select one operator in ${parent}_2$ and use it to replace another random operator in ${parent}_1$, ensuring that the selected operator cannot be the same as the one it is replacing. 
Therefore, our crossover is single-point and uniformly random.
After crossover, we perform additional mutations on the child architecture.

In the \textit{mutation} procedure, we select architectures that are members of $P_{best}$ or that came from the crossover procedure and 
perform 1-edit random mutations on 
their 
internal structures. 
The actual definition for 1-edit change is determined by the search space. 
For example, on NAS-Bench-101, 1-edit mutations include the following:
\begin{itemize}
    \item Swap an existing operator in the cell with another uniformly sampled operator.
    \item Add a new operator to the cell with random connections to other operators.
    \item Remove an existing operator in the cell and all of its incoming/outgoing edges.
    \item Add a new edge between two existing operators in the cell.
    \item Remove an existing edge from the cell.
\end{itemize}
It is also worth mentioning that in our search we allow for more than 1-edit mutations, i.e., we could randomly perform consecutive mutations on an architecture to boost the diversity of the new population $P_{new}$.

\begin{table*}[t!]
\caption{Summary of key hyper-parameters for our EA search algorithm. $P_{init}$ refers to the 
starting population.}
\label{tab-hparams}
\small
\begin{center}
\scalebox{0.9}{
\begin{tabular}{|l|cccc|cccc|cccc|} \hline
   \multirow{2}{*}{\textbf{Method}}                 & \multicolumn{4}{c|}{\textbf{NAS-Bench-101}} & \multicolumn{4}{c|}{\textbf{NAS-Bench-201}} & \multicolumn{4}{c|}{\textbf{NAS-Bench-301}} \\\cline{2-13}
                   & $k$ & $B$ & $|P_{init}|$ & $T$   & $k$ & $B$ & $|P_{init}|$ & $T$    & $k$ & $B$ & $|P_{init}|$ & $T$  \\ 
\hline
Random/Synflow     & 20 & 100 & 100 & 6        & 10 & 20  & 10 & 4     & 20 & 100 & 100 & 7\\ \hline
CL-fine-tune       & 20 & 100 & 50  & 6        & 10 & 10  & 10 & 5     & 20 & 100 & 50  & 7\\
\hline
\end{tabular}
}
\end{center}
\vspace{-6mm}
\end{table*}

We evaluate the performance of each architecture in $P_{new}$ using a given performance predictor $M$. We denote a NAS-Benchmark we can query as $D$ as well as a query budget $B$. Using $D$, we query the ground truth performance of the first $B$ architectures in $P_{new}$ and denote this labelled subset $P_{child}$. We add $P_{child}$ to the existing population $P$ and continue to the next iteration $t+1$. Algorithm~\ref{alg:ea} summarizes our overall search procedure, while Table~\ref{tab-hparams} enumerates our specific hyperparameter settings. 

\subsection{Predictor Components}

Our CL graph encoder considers Transformer~\cite{vaswani2017attention} attention mechanisms with positional embeddings~\cite{chen2020simple, chen2020big} to capture global graph relationships and $k$-GNN~\cite{morris2019weisfeiler} layers to capture local features. 
Starting from operation node embeddings,  
the transformer encoder consists of up to 2 layers with 2 attention heads, whose output we concatenate with a $k$-GNN~\cite{morris2019weisfeiler} to form an overall graph embedding of size $m=$ 128. We generate graph embeddings by aggregating the features of all nodes using the mean operation. An MLP with 4 hidden layers and ReLU projects the graph embedding down to size $p=$ 64 to compute the CL loss. When training the CL encoder, we set $\tau=$ 0.1 and normalize output representations as \cite{khosla2020supervised} suggest. 
Finally, we use a separate MLP head with 5 layers and a hidden size of 200 to make a prediction from the CL embedding. 

Our simple GNN graph encoder baseline consists of 4 or 6 layers with node feature size 64. We apply the same mean aggregation to form a graph embedding, and then use an MLP head with 4 hidden layers and ReLU activation to make a prediction. 

\end{document}